%% file: aaai22.tex
\def\year{2022}\relax
\documentclass[letterpaper]{article} % DO NOT CHANGE THIS
\usepackage{aaai22}  % DO NOT CHANGE THIS
\usepackage{times}  % DO NOT CHANGE THIS
\usepackage{helvet}  % DO NOT CHANGE THIS
\usepackage{courier}  % DO NOT CHANGE THIS
\usepackage[hyphens]{url}  % DO NOT CHANGE THIS
\usepackage{graphicx} % DO NOT CHANGE THIS
\usepackage{natbib}  % DO NOT CHANGE THIS AND DO NOT ADD ANY OPTIONS TO IT
\usepackage{caption} % DO NOT CHANGE THIS AND DO NOT ADD ANY OPTIONS TO IT
\usepackage{lineno}
\usepackage{hyperref}
\usepackage{multirow}
\usepackage[table,xcdraw]{xcolor}

\DeclareCaptionStyle{ruled}{labelfont=normalfont,labelsep=colon,strut=off} % DO NOT CHANGE THIS
\usepackage{algorithm}
\usepackage{algorithmic}
\usepackage{xcolor}
\usepackage{newfloat}
\usepackage{listings}

\floatstyle{ruled}
\newfloat{listing}{tb}{lst}{}
\floatname{listing}{Listing}
\title{NADBenchmarks - a compilation of Benchmark Datasets for Machine Learning Tasks related to Natural Disasters}
\author{
    Adiba Mahbub Proma, %\textsuperscript{\rm 1}
    Md Saiful Islam, %\textsuperscript{\rm 1}
    Stela Ciko, %\textsuperscript{\rm 1}
    Raiyan Abdul Baten, 
    Ehsan Hoque %\textsuperscript{\rm 1}
}
\affiliations{
    University of Rochester, United States
    
}

\usepackage{bibentry}
\begin{document}
\maketitle

\begin{abstract}

\input{abstract}

\end{abstract}
\section{Introduction} 
%Climate change has increased the intensity, frequency, and duration of extreme weather events and natural disasters across the world \cite{portner2022climateIPCC}.  Research shows that the historically extreme rainfall during Hurricane Harvey in North America was 3 to 10 times more likely a consequence of climate change \cite{portner2022climateIPCC} \textcolor{red}{[note, actual citation should be chapter 14]}. This is not a sparse event. Across the world, floods have worsened, wildfires have become more frequent, and heat waves now last longer. These events result in massive displacement, loss of lives and severe economic damage both short term and long term. According to the NOAA National Centers for Environmental Information, the total damage in the USA from natural disasters between 2017 to 2022 have been estimated at \$798.4 billion, with a death count of 4,565 people \cite{noaaGov}. For Hurricane Harvey alone, the average economic damage estimation was about US \$90billion \cite{frame2020economiccostofHarvey}. Therefore, a crucial part of climate change adaptation now includes tackling natural disasters. 
With climate change exacerbating the intensity, frequency, and duration of extreme weather events and natural disasters across the world \cite{portner2022climateIPCC}, rapid advancement is required in its management. Researchers are taking advantage of the huge amount of data available, applying ML for more effective solutions in managing natural disasters \cite{automatedML, yu2018big}. For example, the \href{https://data.noaa.gov/datasetsearch/}{NOAA website}\footnote{https://data.noaa.gov/datasetsearch/} contains carefully curated climate change related data that researchers can use. However, progress is relatively slow. In fields of computer vision and natural language processing, benchmark datasets have played a crucial role in their recent rapid advancement and we believe that developing ML algorithms for natural disaster management can benefit from such rigor. Benchmark datasets are preprocessed, curated datasets for training and testing ML algorithms. These datasets provide scope for standard evaluation, allowing ML communities to quantify their progress and compare their models against each other. Existing review papers focus on the role of big data and machine learning to tackle natural disasters \cite{automatedML, yu2018big, resch2018combining}, but to our knowledge, do not cover the scope of benchmark datasets for natural disasters, and their potential in accelerating research in this domain. 

The objective of this short paper is to explore the state of benchmark datasets for ML tasks related to natural disasters, categorizing the datasets according to the disaster management cycle, which consists of four stages - mitigation, preparedness, responses and recovery \cite{khan2008disaster}. The \textbf{mitigation} phase deals with making long term plans to reduce the effects of a disaster; the \textbf{preparedness} phase focuses on plans for responding to a disaster; \textbf{response} activities include damage assessment and providing post-disaster coordination; and the \textbf{recovery} phase relates to recovering from the damages. For this task, we compile a list of existing benchmark datasets introduced in the past five years, find current gaps in literature, and discuss their implications. To facilitate research in this domain, we propose a web platform - NADBenchmarks - where researchers can search for benchmark datasets related to the topic. Our goal is to increase accessibility for researchers, instead of generating a leaderboard platform since leaderboards often pose the risk of simplifying progress to a singular metric \cite{utilityNLPcriticism}. Our preliminary version of such a platform developed using our compiled list can be found in this \href{https://roc-hci.github.io/NADBenchmarks/}{link}\footnote{https://roc-hci.github.io/NADBenchmarks/}.

%Over the years, researchers have been collecting data on this topic from surveys, Earth observation, climate simulations, and sensor networks \cite{yu2018big}. Recently, citizen-generated data have become a promising data source, especially through crowdsourcing efforts, and social media. Social media has also been used for real-time monitoring of natural disasters. For example, during Hurricane Sandy, before and after tweets of fifty US cities were analyzed to find the per-capita economic damage caused \cite{kryvasheyeu2016rapid}. 

%The objective of this short paper is to explore the state of benchmark datasets for ML tasks related to natural disasters, categorizing the datasets according to the disaster management cycle. For this task, we compile a list of existing benchmark datasets that have been introduced in the past five years. To facilitate research in this domain, we propose a a web platform where researchers can search for benchmark datasets related to the topic. We develop a preliminary version of such a platform using our compiled list, and this can be found at the link. Moreover, we find current gaps in literature and discuss the implications of our findings. This paper is intended to facilitate the search for benchmark datasets for researchers in this domain and provide general directions in finding topics where they can contribute new benchmark datasets. 

%___________________________________IMPORTANT________________________________
\begin{table*}[t]
\centering
\small 
\begin{tabular}{|l|l|l|}
\hline
\multicolumn{1}{|c|}{\textbf{Phases in Disaster Cycle}} & \multicolumn{1}{c|}{\textbf{\begin{tabular}[c]{@{}c@{}}Examples of Benchmark \\ Applications\end{tabular}}} & \multicolumn{1}{c|}{\textbf{Type of ML Tasks}} \\ \hline
\multirow{2}{*}{\begin{tabular}[c]{@{}l@{}}Prevention - Forecasting \\ and prediction\end{tabular}} & Drought forecasting & Multiclass ordinal classification \\ \cline{2-3} 
 & \begin{tabular}[c]{@{}l@{}}Earth surface forecasting; \\ Extreme summer prediction; \\ seasonal cycle prediction\end{tabular} & Video prediction \\ \hline
Preparedness - Early warning & Wildfire spread prediction & Image segmentation \\ \hline
\multirow{4}{*}{\begin{tabular}[c]{@{}l@{}}Preparedness - Monitoring \\ and detection\end{tabular}} & Ground deformation detection & Binary classification \\ \cline{2-3} 
 & Real-time wildfire smoke detection & Binary classification \\ \cline{2-3} 
 & Disaster detection & \begin{tabular}[c]{@{}l@{}}Binary classfication, multiclass classification, \\ multiclass multilabel classification\end{tabular} \\ \cline{2-3} 
 & Flood detection & Binary classification, image segmentation \\ \hline
\multirow{3}{*}{Response - Damage assessment} & Damage severity assessment & \begin{tabular}[c]{@{}l@{}}Multiclass ordinal classification, \\ Semantic segmentation, multilabel \\ classification, multitask learning\end{tabular} \\ \cline{2-3} 
 & Damage assessment of buildings & \begin{tabular}[c]{@{}l@{}}Image segmentation, multiclass \\ ordinal classification\end{tabular} \\ \cline{2-3} 
 & Post Flood Scene Understanding & \begin{tabular}[c]{@{}l@{}}Image classification, semantic segmentation, \\ visual question answering\end{tabular} \\ \hline
\multirow{2}{*}{\begin{tabular}[c]{@{}l@{}}Response - Post-disaster \\ Coordination and Response\end{tabular}} & Assessing informativeness & Binary classification, multitask learning \\ \cline{2-3} 
 & Categorization of humanitarian tasks & Multiclass classification, multitask learning \\ \hline
Recovery & Sentiment analysis & Multiclass multilabel classification \\ \hline
\end{tabular}
\caption{This table summarizes some common ML applications where benchmark datasets have been introduced for natural disasters. A full list can be found in our webpage.}
\label{tab:summary}
\end{table*}

%___________________________________IMPORTANT________________________________

\section{Method} 
\textbf{Search Criteria: }We searched different combinations of the keywords `ML datasets', `natural disaster', and `benchmark' in Google Scholar, ACM digital library and Scopus. To capture the latest works, we collected additional papers from popular ML conferences CVPR, NeurIPS, ICML and ICLR; and the website climatechange.ai \cite{rolnick2022tackling}. We especially targeted papers that introduced datasets for a particular task. For this short review, we focused on papers from the past five years (2017-2022) since we wanted to capture the most recent works as a representation of the current state of the art for this domain. 

 \noindent\textbf{Data Extraction and Curation: }11 criteria related to benchmark dataset characteristics are selected and data is extracted from the papers accordingly. These criteria are - dataset name, application, ML task, natural disaster topic, phase (and subphase) in disaster management cycle, timespan, geological coverage, dataset type, size, and data source. Additionally, the paper title, venue and the year published are also extracted. For space constraints, the raw data is not provided in the paper but Table~\ref{tab:summary} has summary examples and a full list is available on the \href{https://roc-hci.github.io/NADBenchmarks/}{webpage}. Implementation details for the webpage are provided in section 5. 

\section{Review of Benchmark Datasets}
\subsection{Prevention/Mitigation}
Prevention refers to long term planning on how to reduce the risk of natural disasters. This phase can be broken down into two sub-phases - Long-term Risk Assessment and Reduction, which refers to analyzing risk and taking steps to mitigate them; and Forecasting and Predicting, which focuses on methods to predict natural disasters.Two of our reviewed datasets focused on forecasting and prediction problems. EarthNet2021 was presented as a challenge for Earth surface forecasting, extreme weather prediction and seasonal cycle prediction using satellite imagery \cite{requena2020earthnet2021}. DroughtED was introduced for drought prediction, classifying drought into six categories (from no drought to exceptional) using drought and meteorological observations, and spatio-temporal data \cite{minixhofer2021droughted}. 

\subsection{Preparedness}
The goal of the preparedness phase is to plan how to respond to a natural disaster, including detecting its progression, and warning citizens. So, it can be categorized into two sub-phases - Monitoring and Detection, and Early Warning. There is only one dataset on early warning - Next Day Wildfire Spread dataset - which is labelled to predict how far the wildfire would spread, given previous images \cite{huot2021nextwildfire}. Most papers focus on monitoring and detection of disasters from images, but with varying complexity. Three papers introduce benchmarks for detecting types of disasters, and thus defining it as a single-label multiclass classification problem \cite{alam2020deep, said2021active, weber2020detecting}. Recent work shows further improvement, introducing datasets for multiclass multilabel learning - Incidents1M builds on the incidents dataset by adding more labels to the images in the incidents dataset \cite{weber2022incidents1m}, and the MEDIC dataset is introduced as an extension of  \citet{alam2020deep}’s work \cite{alam2021medic}. The community has also started exploring the potential of video datasets - VIDI contains 4,534 video clips with 4,767 labels and provides a baseline for multilabel disaster detection \cite{sesver2022vidi}. 

One of the most common topics in detecting specific natural disasters is  flood monitoring and detection from Earth observation data such as satellites and SARs. MediaEval introduced flood-related challenges for three consecutive years (2017-2020), starting with flood detection, flooded road detection, and flood severity detection \cite{2017mediaEval, 2018mediaEval, 2019mediaEval}. Similarly, Spacenet 8 introduced a benchmark challenge for flood mapping on road segments and buildings \cite{hansch2022spacenet}. A benchmark for flood extent detection is introduced by \citet{gahlot2022curating}, where special focus is given to distinguishing flood and general water bodies. Datasets are also introduced for real-time wildfire smoke detection \cite{2022figlib} and volcanic stage classification \cite{2022hephaestus}. 

\subsection{Response} 
Response deals with tackling the immediate aftermath of the disaster. This includes assessing and estimating the damage caused, and taking steps to aid those affected by the disaster. This phase can be divided into two subphases - Damage Assessment and Post-disaster Coordination and Response. 

Multiple papers have introduced benchmarks for damage classification using social media \cite{zhu2021msnet, mouzannar2018damage}, Earth observation data \cite{gupta2019xbd, rahnemoonfar2021floodnet, chowdhury2022rescuenet, chen2018benchmark} or climate-simulated data \cite{kashinath2019climatenet} with images or in multimodal format (text and images) \cite{nguyen2017damage,mouzannar2018damage}. xBD also introduces the Joint Damage Scale as a unified scale for damage assessment of all natural disasters through satellite imagery \cite{gupta2019xbd}. Some work has also been done in assessing building damage post Hurricane using satellites \cite{chen2018benchmark} and aerial videos \cite{zhu2021msnet}; and for flood scene understanding \cite{rahnemoonfar2021floodnet}. A more comprehensive annotation is provided by RescueNet, consisting of different damage levels for 11 different categories, including debris \cite{chowdhury2022rescuenet}. 
During post-disaster coordination, the limited resources must be allocated properly for rescue and relief operations. Benchmarks created over the years deal with assessing the informativeness of data from social media, and categorizing the type of humanitarian aid required. One of the earlier benchmarks is the CrisisMMD dataset \cite{alam2018crisismmd}, where tweets from seven natural disasters during 2017 were annotated for informativeness as a binary classification problem, and across eight humanitarian categories as a multiclass classification problem. HumAID builds on concepts from CrisisMMD, introducing a larger dataset for humanitarian aid classification, labelling tweets for 19 natural disasters across ten categories \cite{alam2021humaid}.

Publicly available datasets were combined to increase size, and both damage severity and humanitarian aid classification were introduced for multilabel classification \cite{alam2020deep}, and then extended to CrisisBench \cite{alam2021crisisbench}. Building upon CrisisBench, the MEDIC dataset provides a benchmark for multitask learning for damage severity and humanitarian aid classification \cite{alam2021medic}. 

\subsection{Recovery}
The last stage is the recovery phase which focuses on aiding people to get their lives back to normalcy. We did not find a paper that was directly related, but we could loosely classify image-sentiment dataset into this category \cite{hassan2022visual}. Crowdworkers label images of natural disasters according to their sentiments on seeing the image,  creating a benchmark for multilabel sentiment classification for natural disasters. In the future, this kind of benchmark can be useful for analyzing the long-term psychological effects on victims of natural disasters. 

\begin{figure}[t!]
\resizebox{0.45\textwidth}{!}{\includegraphics{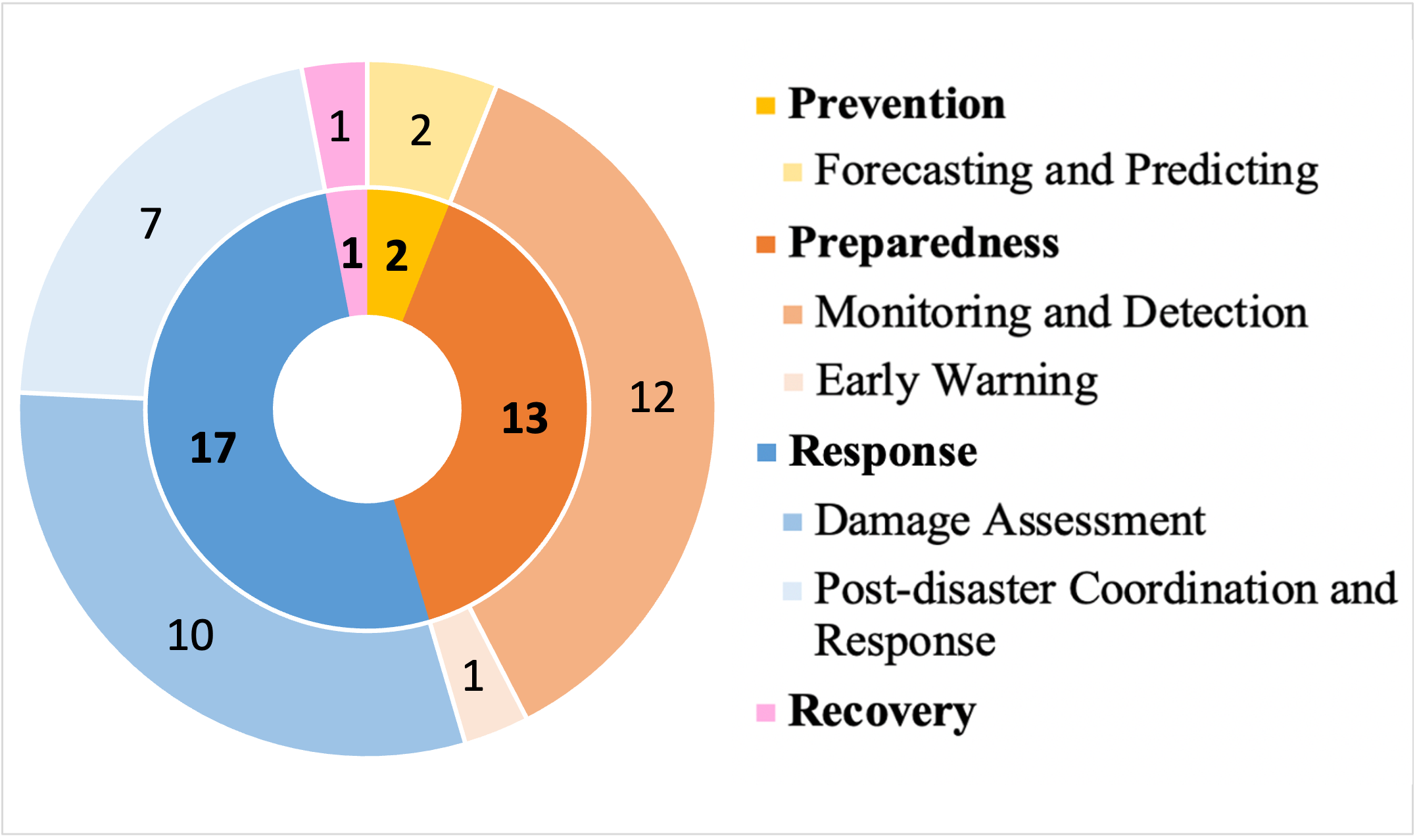}}
\caption{Pie-chart showing the number of papers found for each of the disaster management phases and subphases.}
\label{fig:doublepie}
\end{figure}

%fix the image
\begin{figure*}[h]
\begin{center}
\includegraphics[width=0.97\linewidth]{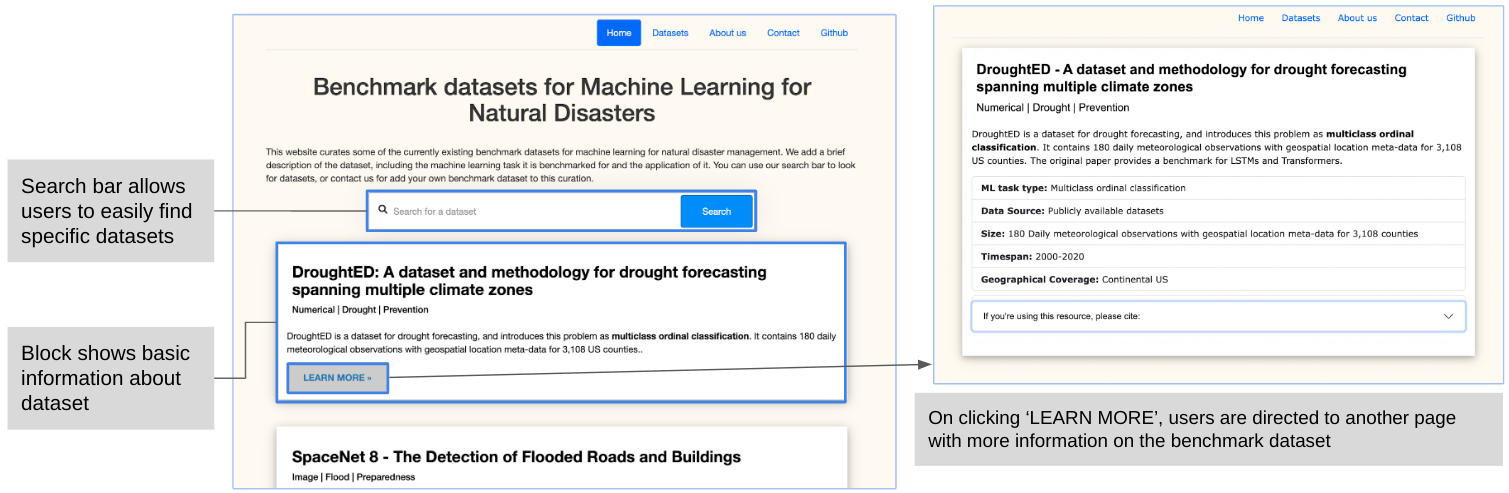}
\caption{Our interface with key features highlighted.}
\label{fig:interface}
\end{center}
\end{figure*}
%----------------------------------------------------------------------
\section{Trends and research gaps} 
\textbf{More datasets are needed for Prevention and Recovery Phase.}
According to our review, most benchmark datasets have been introduced for the preparedness and the response phase, as shown in figure~\ref{fig:doublepie}. Only two papers focused on prevention, and we could loosely categorize one paper for recovery. However, the lack of benchmark datasets must not be confused with the lack of applications of ML. Multiple works have been published on risk assessment during floods \cite{skakun2014flood}, earthquakes \cite{wilson2016rapid, ehrlich2013optical} and hurricanes \cite{ehrlich2013optical}. Similarly, ML has been used for change detection in the recovery phase and also to predict the economic consequences of such outlier events \cite{gurrapu2021deepag}. Various data sources have been used so far, including satellites, crowdsourced data, financial records and call logs \cite{skakun2014flood, wilson2016rapid, ehrlich2013optical, gurrapu2021deepag}. These examples show that there is utility in generating benchmark datasets for these phases. Some potential ideas for benchmark datasets include datasets for risk assessment, evaluating or predicting recovery process, predicting displacement patterns post-disaster, and so on. 

\textbf{More diversification is needed in data type and data sources.} 
Currently, there is disproportionate use of social media and Earth observation data (satellites, UAV and drones) as data sources. This can be attributed to their increased availability, accessibility and their relative inexpensiveness. Many people turn to social media for help during a crisis, thus contributing to a large pool of resources, ideal for data mining. Organizations also make it easy for researchers to use their data. Twitter, for example, has an API that allows researchers to access tweets through querying hashtags. Moreover, remote sensing makes it easy to collect images before, during and after a natural disaster without putting further lives at risk. While these are very important sources, they still have their limitations. For instance, occlusion due to clouds or smoke is a common issue for satellite imagery. This is especially exacerbated by the fact that in case of some natural disasters such as hurricanes or wildfire, clouds and smoke are inevitable. UAVs and drones can often miss specific angles, and thus miss crucial information. Moreover, although social media provides a large resource, they are often noisy and require a lot of preprocessing. 

Most benchmarks reviewed in the paper are in image format. In fact, social media and Earth observation data are usually in image format and considering the rapid improvement in machine vision algorithms, it makes sense that most problems in this domain are presented as vision problems. However, this increases the research communities’ risk of running into limitations and missing out on the advantages of other data types and sources. Commercial data sources such as cell networks, call logs, and financial records can be good resources for damage assessment tasks and mobility prediction \cite{smallwood2022mobile}. Multimodal approaches that include audio and text data could improve the performance of existing algorithms. 

%\subsection{Generalizability vs domain-specific benchmarks} 
%Most benchmark datasets were focused on general disaster detection. While generalizability of models is important, it`s also important that focus is put on creating domain-specific benchmarks because each natural disaster comes with its own challenges, which are not addressed in generalized models. 

\textbf{Benchmarks for multitask learning problems.}
Multitask learning (MTL) has been a promising method for achieving generalizability and improving efficiency, and researchers are just starting to explore the suitability of MTL for climate models \cite{gonccalves2015multitask}. MTL models work under the assumption that the tasks are related to one another. However, trying to build a unified approach for unrelated tasks can be detrimental to performance and there is ongoing research on what tasks learn better together \cite{standley2020tasks}. Some papers have started discussing the scope of building MTL benchmark datasets for this domain. So far, we were able to find only one paper on benchmark for MTL \cite{alam2021medic}, but we can expect more in the future.

\section{Interface implementation}
Taking inspiration from current benchmark data curation websites such as CrisisNLP (datasets for crisis management), GEM (datasets for natural language generation), and paperswithcode (datasets for vision tasks), we are building a web platform to increase data accessibility for researchers in this domain. Currently, our working prototype consists of information on the benchmark datasets reviewed in this paper. Users can scroll through our list of datasets, each of which consists of a short description explaining the ML task and the topic of the benchmark. Interested researchers can click the `Learn More' button for more information, as shown in figure~\ref{fig:interface}. Currently, we display the data extracted using the 11 criteria. In the future, we aim to add more features, including information on annotation methods, ML models the benchmark dataset has been tested on, labels, data distribution and accuracy metrics.

%CrisisNLP is a website where they host datasets that researchers working on crisis informatics can easily use. GEM2 is a multilingual NLG benchmarking, and that is also hosted in a website, where data description and visualization of the datasets is shown. These websites make finding appropriate data and setting up the environment easy. We aim to build an interactive website where researchers can learn about the data. Taking inspiration from CrisisNLP and GEM2, we would like to add data description, some examples of raw data and some visualizations of data distribution to our interface. The full model of our interface is still a work in progress, but some early design prototypes are shown in figure 2. Currently our working prototype consists of the 11 criteria that we extracted from the papers. It is in tabulated format for easy understanding, and contains the link to the original paper, and the dataset. 

\section{Discussion and conclusion} 
%In this paper, we briefly review existing benchmark datasets established for ML applications for natural disasters, categorizing them according to the disaster management cycle. While reviewing the papers, we focused on the applications and ML tasks introduced by the original paper. This was appropriate for our analysis since our goal was to facilitate the task of searching for datasets for interdisciplinary researchers, who may not be aware of the different sources of datasets available to them. Researchers can quickly look through the list and get a general idea on which aspects, benchmark datasets have been introduced and where there is scope for further research. 

In this paper, we briefly review existing benchmark datasets for natural disasters, categorizing them according to the disaster management cycle. We focused on the applications and ML tasks introduced by the original paper because our goal was to facilitate the task of searching for datasets for interdisciplinary researchers. 
%, who may not be aware of the different sources available to them. 

%Researchers can quickly look through the list and get a general idea of where there is scope for further research.

However, there are other characteristics of benchmark datasets that are also of equal importance - evaluation metrics, annotation methods, and data distribution \cite{datastatementsNLP, olson2017pmlb}. This is reflected in current vision and NLP benchmark websites - GEM aims at improving evaluation strategies for the NLG community, and paperswithcode ranks papers for vision tasks in terms of their accuracy \cite{GEM, vision_paperswithcode}. A comprehensive analysis of these characteristics were not conducted for our review but can be included in the future. Further research can be done to determine the utility of leaderboards for this domain, and more holistic metrics to evaluate models can be proposed. For example, “data cards” have been introduced in NLP as a more holistic metric which takes bias into account \cite{datastatementsNLP}. Future directions can also focus on generating such data cards for ML for environmental science-related tasks. Despite the limitations, there is scope for inclusion of such information on our webpage. Moreover, as new datasets are introduced, we plan to closely monitor and update our website regularly. We hope the community would contribute to this curation by submitting their datasets. Our goal for future work involves creating a more comprehensive list of benchmark datasets, increasing the scope of information for the datasets, and providing general analytics to inform researchers about the current state of the art. 

\section*{Acknowledgement}
This research was supported by funding from the Goergen Institute for Data Science. 
\bibliography{aaai22}
\end{document}

%% file: abstract.tex
Climate change has increased the intensity, frequency, and duration of extreme weather events and natural disasters across the world. While the increased data on natural disasters improves the scope of machine learning (ML) in this field, progress is relatively slow. One bottleneck is the lack of benchmark datasets that would allow ML researchers to quantify their progress against a standard metric. The objective of this short paper is to explore the state of benchmark datasets for ML tasks related to natural disasters, categorizing them according to the disaster management cycle. We compile a list of existing benchmark datasets introduced in the past five years. We propose a web platform - NADBenchmarks - where researchers can search for benchmark datasets for natural disasters, and we develop a preliminary version of such a platform using our compiled list. This paper is intended to aid researchers in finding benchmark datasets to train their ML models on, and provide general directions for topics where they can contribute new benchmark datasets.